\documentclass[journal]{IEEEtran}
\usepackage{cite} 
\usepackage{hyperref} 
\usepackage{amssymb}

\usepackage{amsmath}
\newtheorem{theorem}{Theorem}
\newtheorem{lemma}{Lemma}
\newtheorem{definition}{Definition}
\newtheorem{corollary}{Corollary}
\usepackage{lineno}
\usepackage{algorithm} 
\usepackage{algorithmic}

\usepackage{hyperref}  
\usepackage{pgfplots}
\pgfplotsset{compat=1.5}
\pgfplotsset{every axis/.append style={thick}}

\begin{document}
\title{Unlimited Budget Analysis of Randomised Search Heuristics Performance}
\author{Jun He and Thomas Jansen and Christine Zarges
\thanks{Manuscript received xx xx xxxx}
\thanks{This work was particially supported by EPSRC under Grant No. EP/I009809/1 (He) . } 
\thanks{Jun He is with the School of Science and Technology, Nottingham Trent University, Nottingham NG11 8NS, U.K. Email: jun.he@ntu.ac.uk. Thomas Jansen and Christine Zarges are with Department of Computer Science, Aberystwyth University Aberystwyth, SY23 3DB, UK. Email: \{t.jansen,c.zarges\}@aber.ac.uk}
\thanks{A preliminary poster version was published in  Proceedings Companion of GECCO 2019~\cite{he2019unlimited}.}} 
\maketitle

\begin{abstract}
The performance of randomised search heuristics is often measured by either the fitness value or approximation error of solutions at the end of running. This is common practice in computational simulation. Theoretical performance analysis of these algorithms is a rapidly growing and developing field. Current work focuses on the performance within pre-defined computational steps (called fixed budget). However,   traditional analysis approaches such as drift  analysis cannot be easily applied to the fixed  budget  performance.  Thus, it is  necessary to develop new approaches. 
This paper introduces a novel analytical approach, called unlimited budget analysis, to evaluating the approximation error or fitness value  after arbitrary computational steps. Its novelty  is on bounding the expected approximation error, rather than fitness value.  
To demonstrate its applicability, several case studies have been conducted in this paper. For random local search and (1+1) evolutionary algorithm on linear functions, good bounds are obtained, although the analysis of linear functions is hard in fixed budget setting. For   (1+1) evolutionary algorithm  on LeadingOnes, bounds obtained from fixed budget performance are extended to  arbitrary computational steps. Furthermore, unlimited budget analysis can be applied to algorithm performance comparison. For  (1+1) evolutionary algorithm on linear functions, its performance under different mutation rates is compared and the optimal rate is identified. For  (1+1) evolutionary algorithm and simulated annealing on the Zigzag function, their performance is compared and simulated annealing may generate slightly better solutions.  These case studies demonstrate unlimited budget analysis is a useful tool of bounding the approximation error or fitness value  after arbitrary computational steps. 
\end{abstract}


\IEEEpeerreviewmaketitle
 
\section{Introduction}
\label{sec-introduction}

Randomised search heuristics (RSH), such as simulated annealing (SA) and evolutionary algorithms (EAs), are general purpose search algorithms  inspired by nature paradigms.   These algorithms share several common features like randomness, iteration, heuristics and search.  An important application domain for RSH is optimisation where one looks for solutions to maximising or minimising  some   functions.

For RSH algorithms in optimisation, there is a growing body of theoretical work that provides insights into how and why RSH works or fails. In particular in the area of EAs such analyses have been concentrated on the aspect of runtime, analysing how long an EA needs to find an optimal solution or a solution with a defined approximation ratio~\cite{jansen2013analyzing}.

An alternative perspective is to consider solution quality achieved by RSH at the end of running. This is  common practice in computer simulation. The quality of a solution can be measured by its function value or  approximation error. 
The two measures are essentially equivalent because the approximation error equals to the difference between the fitness value of a solution and the optimal fitness value.

Fixed budget performance~\cite{jansen2014performance}, whose target aims to derive results about the expected function value achieved by RSH within a pre-defined number of computational steps. But a direct estimation of the expected fitness value is not easy.  Traditional analysis approaches such as drift  analysis cannot be easily applied to fixed budget performance. Even for (1+1) EA on  linear functions, it is   hard to derive lower or upper bound on the fitness value within fixed budget setting~\cite{lengler2015fixed,vinokurov2019fixed}.  Therefore, it is  necessary to develop new approaches. 

This paper introduces a novel analytical approach for  evaluating the expected fitness value and approximation error after an arbitrary number of computational steps. So, it is named unlimited budget analysis. Unlike fixed budget setting,  the   fitness value is not estimated directly; instead, the  approximation error is bounded first, then the fitness value. Our research hypothesis is that the expected approximation error or fitness value can be derived from the convergence rate. Its idea is build upon Rudolph's~\cite{rudolph1997convergence} early work on the convergence rate. Let $e^{[t]}$ denote the approximation error at the $t$th step. Under the condition $\lambda_l e^{[t]} \le e^{[t+1]} \le \lambda_u e^{[t]}$, we have geometrically fast convergence rate: $e^{[0]} (\lambda_l)^t \le e^{[t]} \le e^{[0]} (\lambda_u)^t$.

The structure and contributions of this paper are summarised as follows: 
\begin{enumerate}
    \item Section~\ref{sec-unlimitedBudgetAnalysis} presents the framework of unlimited budget analysis for bounding the expected fitness value via approximation error achieved by RSH after arbitrary computational steps.
    \item Sections~\ref{sec-caseStudies} conducts case studies of local random search and (1+1) EA on LeadingOnes and linear functions, and derives lower and upper bounds on the expected approximation error and fitness value.
    \item Section~\ref{sec-comparison} conducts two case studies. One is to compare the performance of (1+1) EA with different mutation rates on linear functions. The other is to compare the performance of (1+1) EA and simulated annealing on Zigzag functions.     
\end{enumerate}

\section{Related work}
\label{sec-relatedWork}
Current work on performance analysis of RSH is classified into two types according to  the number of computational steps.
\begin{enumerate} 
    \item \textbf{fixed budget setting:}  analysis of the expected fitness value or approximation error restricted to a fixed number of computational steps~\cite{jansen2014performance};
    \item \textbf{unlimited budget setting:} analysis of the expected fitness value or  approximation error for an arbitrary number of computational steps~\cite{he2016analytic,he2018theoretical}.
\end{enumerate}  

The two types bear obvious similarity. However, a number of significant differences exist.
The first difference is fixed against unlimited budget. 
The fixed budget setting considered a fixed number of computational steps $b$ (called budget)~\cite{jansen2014performance}. Results hold for any number of steps $t \leq b$ but may not for $t >b$. In principle, $b$ can be set to arbitrary values but it is recommended to concentrate on budgets that are bounded above by the expected runtime of RSH~\cite{jansen2014performance}. The unlimited budget setting removes the restriction of fixed budget $b$ and investigates any $t \in [0, +\infty)$. Let's $f^{[t]}$ denote the expected fitness value at the $t$-the step. Fixed budget setting aims to approximate $f^{[t]}$ within $[0,b]$, while unlimited budget setting seeks to an approximation of $f^{[t]}$ for $t \in [0, +\infty)$.  

The second difference is analysis methods. In fixed budget setting,  the goal is to bound the fitness value directly. This bounding is often problem-specific~\cite{jansen2014performance}. There are attempts of applying  runtime analysis techniques, such as  Chebyshev's inequality, Chernoff bounds and drift analysis, to fixed budget analysis~\cite{doerr2013method}.  
Nevertheless, as shown in the work~\cite{lengler2015fixed,vinokurov2019fixed}, traditional approaches like drift analysis cannot easily be extended to the fixed budget performance, even for linear functions. Therefore, it is necessary to develop new approaches. 

In unlimited budget setting, the primary goal is to bound the approximation error. Conversion from a bound on the approximation error to a bound on the fitness value is straightforward. A general Markov chain approach was proposed for estimating the approximation error of EAs~\cite{he2018theoretical}. In theory, exact expressions of the approximation error were also obtained for elitist EAs in~\cite{he2016analytic,he2018theoretical}. Furthermore, methods for bounding the approximation error were developed in~\cite{he2018theoretical}.

Up today, most existing work on performance analysis of RSH is within fixed budget setting.  
Jansen and Zarges~\cite{jansen2014reevaluating} proved immune-inspired hyper-mutations
 outperform random local search  on several selected problems from fixed budget perspective.
Lengler and Spooner \cite{lengler2015fixed} analysed the fixed budget performance of  (1+1) EA on linear functions. They adopted two methods,  drift analysis and differential equation plus Chebyshev's
inequality, to derive  general results for linear functions and tight fixed budget results for
the OneMax function. 
Nallaperuma, Neumann and Sudholt\cite{nallaperuma2017expected} applied the fixed budget analysis to the well-known travelling salesperson problem. They bounded the expected fitness gain of random local search,  (1+1) EA and $(1+\lambda)$ EA  within a fixed budget. 
Lissovoi et al.~\cite{lissovoi2017theoretical}  discussed the choice of bet-and-run parameters to maximise expected fitness within a fixed budget.

Recently, Vinokurov et al. \cite{vinokurov2019fixed} analysed  (1+1) EA with resampling on the OneMax and BinVal problems and obtained some improved fixed budget results on them. 
Doerr et al. \cite{doerr2019optimal} compared drift-maximisation with random local search  within fixed budget setting. In fixed budget setting, they considered the fitness distance to the optimum, that is the approximation error in our paper.

On the side of unlimited budget setting, He~\cite{he2016analytic}  gave an exact error expression for (1+1) strictly elitist EAs.  He and Lin~\cite{he2016average} defined the average convergence rate and proved that the error of  the convergent EA  modelled by a homogeneous Markov chain  is bounded by an exponential function  of the number of steps. He et al.~\cite{he2018theoretical} proposed a theory of error analysis based on Markov chain theory. This paper is a further development  and application along this  direction.

Finally, we note that the spirit of the approach presented in this paper is similar to multiplicative drift analysis \cite{doerr2012multiplicative}, but their goals are completely different: fitness value against runtime. Multiplicative drift was also used to derive results in the fixed budget setting on the fitness value \cite{lengler2015fixed}.

\section{Unlimited budget analysis}
\label{sec-unlimitedBudgetAnalysis}
 
\subsection{Randomised search heuristics and mathematical models}
This paper considers the problem of maximising a function,  
\begin{eqnarray}
\max f(x), &\mbox{ subject to } x \in \mathcal{S},
\end{eqnarray}
where $f(x): \mathcal{S} \to \mathbb{R}$ is called a fitness function and $\mathcal{S}$ is  its definition domain.  $\mathcal{S}$ is a finite  set or a closed set in $\mathbb{R}^n$. Denote the maximal fitness value $f^*=\max\{ f(x); x \in \mathcal{S}\}$ and optimal solution set $X^*=\{x \mid f(x)=f^*\}$. 

RSH, described in Algorithm~\ref{alg1},  is often applied to   the above optimisation problem.  An individual $x \in \mathcal{S}$ is a single solution and a population $X  \subset \mathcal{S}$  is a collection of individuals.  

\begin{algorithm}
\caption{Randomised search heuristics}  \label{alg1}
\begin{algorithmic}[1]
\STATE generation counter $t\leftarrow 0$,
\STATE  population $X^{[0]} \leftarrow$ initialise  a population  of solutions subject to a  probability distribution  $\Pr(X^{[0]})$ on $\mathcal{S}$;   
\WHILE{stopping criterion is not satisfied}
\STATE  population $X^{[t+1]} \leftarrow$  generate a new population of solutions subject to a conditional transition probability $\Pr(X^{[t+1]}  \mid X^{[0]} , \cdots, X^{[t]} )$; 
\STATE generation counter $t \leftarrow t+1$;
\ENDWHILE
\end{algorithmic}
\end{algorithm}

\begin{definition} 
The fitness value of population $X^{[t]}$ is  $f(X^{[t]})=\max \{f(x); x \in X^{[t]} \}$ and its expected value is denoted by $f^{[t]}=\mathbb{E}[f(X^{[t]})]$.  
 \end{definition}

Besides the fitness value,  the approximation  error is an alternative measure of solution quality~\cite{he2016analytic,he2018theoretical}.
\begin{definition}  
The {approximation error} of $X^{[t]}$ is   $e(X^{[t]})  =|f(X^{[t]})-f^*|$ and its  expected value is denoted by 
$e^{[t]}=\mathbb{E}[e(X^{[t]})]$.
\end{definition}

Both $f^{[t]}$ and $e^{[t]}$ are functions of $t$. They  depend on $X^{[0]}$ although this dependency is not explicitly expressed.

\begin{definition}
RSH is called  {convergent in mean} if  for any initial population $X^{[0]}$, 
\begin{align}
    \lim_{t \to \infty} e^{[t]}=0, \mbox{ i.e., } \lim_{t \to \infty} f^{[t]}=f^*.
\end{align}
\end{definition}

\begin{definition}
RSH is called {elitist} if $e(X^{[t+1]}) \le  e(X^{[t]})$ for any $t$, or {strictly elitist} if  $e(X^{[t+1]}) < e(X^{[t]})$ for any $t$.  
\end{definition}

Two mathematical models are often used in the study of RSH, which provide necessary mathematical tools. 
\begin{enumerate}
\item \textbf{Supermartingales.}
RSH is modelled by a  {supermartingale} if for any $t$,   $\mathbb{E} [e(X^{[t+1]})]\le   e(X^{[t]}) $. This means the fitness value increases in mean.
Elitist RSH is a supermartingale because $e(X^{[t+1]})\le e(X^{[t]})$. Non-elitist RSH may be a supermartingale too because the condition $\mathbb{E} [e(X^{[t+1]})]\le e(X^{[t]})$ does not require   $e(X^{[t+1]})\le e(X^{[t]})$.  

\item \textbf{Markov chains.}
RSH is   modelled by a  {Markov chain } if for any $t$, the conditional probability $\Pr(X^{[t+1]} \mid X^{[0]}, \cdots, X^{[t]})=\Pr(X^{[t+1]} \mid  X^{[t]})$. This means the state of $X^{[t+1]}$ only depends upon $X^{[t]}$, but not on history.
\end{enumerate}

\begin{lemma}
If the error sequence $\{e(X^{[t]}); t=0,1, \cdots \}$ is a supermartingale and converges in mean, then  ${e^{[t+1]}}/{e^{[t]}}\le \lambda$ for some positive $\lambda<1$. 
\end{lemma}
\begin{IEEEproof}
The sequence $\{e^{[t]}\}$ is a supermartingale, so it always converges to a non-negative constant. If $\lambda=1$, then it does not converge in mean. Thus $\lambda<1$.
\end{IEEEproof}
 
\subsection{Unlimited budget analysis}  
\label{secAnalysis}
Given a sequence $\{X^{[t]};t=0,1, \cdots\}$, we  aim to find  a  bound (lower or upper) on the fitness value $f^{[t]}$, which is a function of $t$  satisfying two conditions: 
\begin{enumerate}
\item the bound holds for any $t \in [0, +\infty)$;  
\item the bound   converges to $f^*$ if    $\lim_{t\to +\infty} f^{[t]} =f^*$. 
\end{enumerate} 
These two requirements do not exist in fixed budget setting.

Because $f^{[t]} =|f^*-e^{[t]}|$,  the above task  is equivalent to finding  a  bound (lower or upper) on the approximation error $e^{[t]}$, which is a function of $t$ satisfying two conditions: 
\begin{enumerate}
\item the bound holds for any $t \in [0, +\infty)$;  
\item the bound   converges to $0$  if    $\lim_{t\to +\infty} e^{[t]} =0$. 
\end{enumerate}  

Unlimited budget analysis  first derives a bound on  $e^{[t]}$, then a bound on $f^{[t]}$.   This  method is different from  fixed budget analysis, which  estimates a  bound on  $f^{[t]}$ directly without considering $e^{[t]}$.

The purpose of this paper is to seek a bound represented by an exponential function such that $e^{[t]} \le e^{[0]} \lambda^t$ for an upper bound or $e^{[t]}\ge e^{[0]} \lambda^t$ for a lower bound. Once a bound on $e^{[t]}$ is obtained, it is straightforward to derive a bound on $f^{[t]}$. 

One method of bounding $e^{[t]}$ is built upon the convergence rate~\cite{he2016average,he2018theoretical} of the sequence $\{e^{[t]}; t=0,1, \cdots \}$. 
\begin{definition}
Given a sequence $\{e^{[t]}; t=0,1, \cdots \}$,
its  convergence rate   at the $t$-th generation is
\begin{equation}
 r^{[t]} =\left\{
 \begin{array}{lll}
 1-\frac{e^{[t+1]}}{e^{[t]}},     & \mbox{if } e^{[t]} \neq 0, \\
  0,    & \mbox{otherwise.} 
 \end{array}   
 \right.
\end{equation}
Its average (geometric) convergence rate for $t$ generations is 
\begin{align}
\label{equAverageRate}
   R^{[t]}=\left\{
 \begin{array}{lll}
 1- \left(  \frac{e^{[t]}}{e^{[0]}} \right)^{1/t},     & \mbox{if } e^{[0]} \neq 0, \\
  0,    & \mbox{otherwise.} 
 \end{array} 
 \right.
\end{align} 
\end{definition}
In the above definition, the convergence rate is normalised to the range $(-\infty, 1]$ so that the convergence rate is understood as the convergent speed.  The larger the convergence rate is, the faster  $e^{[t]}$ converges to $0$. A negative value of the convergence rate means that $X^{[t]}$ moves away from the optimum $X^*$.

It is straightforward to bound  $e^{[t]}$ and $f^{[t]}$ from the convergence rate of $e^{[t]}$.  The theorem below  originates from~\cite[Theorem 2]{rudolph1997convergence} and is revised in~\cite{he2018theoretical}.

\begin{theorem}
\label{theoremBasic}
Given an error  sequence $\{e^{[t]}; t=0,1, \cdots \}$, if    there exist some $\lambda_l>0, \lambda_u>0$, and for any $t \in [0, +\infty)$, $\lambda_l \le {e^{[t+1]}}/{e^{[t]}}\le \lambda_u$, then 
\begin{align}
e^{[0]} (\lambda_l)^t \le &e^{[t]} \le e^{[0]} (\lambda_u)^t,   \\ 
f^*-e^{[0]} (\lambda_u)^t \le &f^{[t]} \le f^*-e^{[0]} (\lambda_l)^t.
\end{align} 
\end{theorem}

\begin{IEEEproof}
It is sufficient to prove the upper bound in the first claim.
From the condition ${e^{[t+1]}}/{e^{[t]}}\le \lambda_u$, we get ${e^{[t+1]}}\le {e^{[t]}} \lambda_u $ and then ${e^{[t]}}\le {e^{[0]}} (\lambda_u)^t $.  
\end{IEEEproof}

If RSH  is modelled by a  Markov chain, we can estimate  ${e^{[t+1]}}/{e^{[t]}}$ from one-step error change. For the sake of analysis, the definition domain $\mathcal{S}$  is assumed to be a finite set.  

\begin{definition}
The average of  {error change}  at $X^{[t]}=X$ is  
\begin{align} 
    \Delta  e(X^{[t]}) =   \mathbb{E}[e(X^{[t]})- e(X^{[t+1]}) \mid X^{[t]}=X].
\end{align}    
The average of  {error change} at the $t$th generation is 
\begin{align} 
    \Delta  e^{[t]} = \mathbb{E}[ \mathbb{E}[e(X^{[t]})- e(X^{[t+1]}) \mid X^{[t]}]].
\end{align}  
The ratio of  {error change} at  $X^{[t]}$  is   
${ \Delta  e(X^{[t]})}/{e(X^{[t]})}
$, which equals to the convergence rate  at $X^{[t]}$.
The ratio of {error change} at the $t$th generation is  
$ {  \Delta  e^{[t]}}/{e^{[t]}}
$.
\end{definition}

Theorem~\ref{theoremOneStep} provides the range of $e^{[t]}$ based on one-step error change.

\begin{theorem}\label{theoremOneStep}
Assume that  the sequence  $\{X^{[t]}; t=0,1, \cdots \}$ is a {Markov chain}  on  a finite state set $S$. Let
\begin{align}
\label{equOneStepdelta}
 \delta_{\min}  =  \inf_{t }  \min_{X:  X \cap X^* =\emptyset}  \frac{ \Delta  e(X^{[t]})}{e(X^{[t]}=X)}, \\
 \delta_{\max}  =  \sup_{t } \max_{X:  X \cap X^* =\emptyset}  \frac{ \Delta  e(X^{[t]})}{e(X^{[t]}=X)}.
\end{align}   Then  
\begin{align}   e^{[0]} \left(1- \delta_{\max}\right)^t \le &e^{[t]} \le e^{[0]} \left(1- \delta_{\min}\right)^t.
\\f^*-e^{[0]} \left(1- \delta_{\min}\right)^t\le &f^{[t]} \le f^*-e^{[0]} \left(1- \delta_{\max}\right)^t.
\label{equGeneralBounds}
 \end{align}
\end{theorem}

\begin{IEEEproof}
We only prove the upper bound in the first claim.  From the definition of $\delta_{\min}$, we have
\begin{align}
\frac{e^{[t+1]}}{e^{[t]}} =1-\frac{\Delta e^{[t]}}{e^{[t]}} \le 1- \delta_{\min}.
\end{align}
Then we get $e^{[t]} \le {e^{[0]}}\left(1- \delta_{\min} \right)^t.
$ 
\end{IEEEproof}

The main task in this paper is to estimate $\delta_{\min}$ and $\delta_{\max}$.
For an elitist RSH algorithm, $\delta_{\min}$ and $\delta_{\max}$ correspond to the minimum and maximal values of the ratio of error change between two fitness levels, but do not depend on the number of fitness levels. Thus, lower and upper bounds (\ref{equGeneralBounds}) are not related to the number of fitness levels. This is completely from runtime.

It is possible to improve lower and upper bounds   on $e^{[t]}$ using  multi-step  error change.  For the sake of illustration,    only  two-step error change is presented here.  

\begin{definition}
The average of  {error change} in two generations  at $X^{[t]}=X$ is  
\begin{align} 
    \Delta''  e(X^{[t]})   =   \mathbb{E}[e(X^{[t]})- e(X^{[t+2]}) \mid X^{[t]}=X].
\end{align}    
The average of  {error change} in two generations  at the $t$th generation is   
\begin{align*} 
    \Delta''  e^{[t]}  = \mathbb{E}[ \mathbb{E}[e(X^{[t]})- e(X^{[t+2]}) \mid X^{[t]}]] =e^{[t]} - e^{[t+2]}.
\end{align*} 
\end{definition}

\begin{theorem}\label{theoremTwoStep}
Assume that the sequence  $\{X^{[t]}; t=0,1, \cdots \}$ is a {  Markov chain} on a finite set $\mathcal{S}$. 
 Let
\begin{align}
\label{equTwoStepdelta}
 \delta''_{\min}  =  \inf_{t }  \min_{X:  X \cap X^* =\emptyset}  \frac{ \Delta''  e(X^{[t]})}{e(X^{[t]}=X)}, \\
 \delta''_{\max}  =  \sup_{t } \max_{X:  X \cap X^* =\emptyset}  \frac{ \Delta''  e(X^{[t]})}{e(X^{[t]}=X)}.
\end{align}   Then  
\begin{align}   e^{[0]} \left(1- \delta''_{\max}\right)^t \le e^{[2t]} \le e^{[0]} \left(1- \delta''_{\min}\right)^t.
 \end{align} 
\end{theorem}

\begin{IEEEproof}
It is sufficient to prove the upper bound.
Since
\begin{align}
\frac{\Delta'' e^{[t]}}{e^{[t]}} \ge   \delta''_{\min} 
\end{align}
we get $
  e^{[2t]} \le e^{[0]} \left(1- \delta''_{\min}\right)^t.
$
\end{IEEEproof}
 
Using two-step error change is more complex than using one-step, but a potential benefit is a tighter bound. This is proven in the following theorem.

\begin{theorem}\label{theoremTwoStep2}
Assume that the sequence  $\{X^{[t]}; t=0,1, \cdots \}$ is a {Markov chain} on a finite set $\mathcal{S}$. Then   
\begin{align}
&1-\delta''_{\min}
\le \left(1-\delta_{\min} \right)^2,\\ 
&1-\delta''_{\max}
\ge \left(1-\delta_{\max} \right)^2.
\end{align} 
\end{theorem}

\begin{IEEEproof}
We only prove the fist conclusion because the second one can be proven in a similar way. Without loss of generality, denote
\begin{align}
X^{\sharp}  =\arg  \inf_{t } \min_{X:  X \cap X^* =\emptyset} \frac{\Delta'' e(X^{[t]})}{ e(X^{[t]}=X)}.
\end{align}
We get for $X^{\sharp}$ 
\begin{align*}  
 &1-   \frac{\Delta'' e(X^{[t]})}{ e(X^{[t]}=X^{\sharp})}  = \frac{\mathbb{E}[e(X^{[t+2]})]}{ e(X^{[t]}=X^\sharp)}\nonumber\\
 =&\frac{\mathbb{E}[e(X^{[t+1]})]}{ e(X^{[t]}=X^{\sharp})} \times \frac{\mathbb{E}[e(X^{[t+2]})]}{\mathbb{E}[ e(X^{[t+1]})]}    \\
\le&  \left(1- \frac{\Delta e(X^{[t]})}{e(X^{[t]}=X^{\sharp})}\right) \left(1- \frac{\Delta e^{[t+1]}}{e^{[t+1]}}\right) \\
\le& \left(1-\inf_{t } \min_{X:  X \cap X^* =\emptyset} \frac{\Delta e(X^{[t]})}{ e(X^{[t]}=X)} \right)^2,  
\end{align*} 
then we come to the first conclusion.
\end{IEEEproof}


At the end, we must mention that the convergence rate can be used to evaluate the fixed budget performance  too.

\begin{corollary}\label{corollary}
Given an error  sequence $\{e^{[t]}; t=0,1, \cdots \}$ and an integer $b$, if    there exist some $\lambda_l>0, \lambda_u>0$, and for any $t \in [0, b)$, $\lambda_l \le {e^{[t+1]}}/{e^{[t]}}\le \lambda_u$, then for any $t\le b$, 
\begin{align}
e^{[0]} (\lambda_l)^t \le &e^{[t]} \le e^{[0]} (\lambda_u)^t,   \\ 
f^*-e^{[0]} (\lambda_u)^t \le &f^{[t]} \le f^*-e^{[0]} (\lambda_l)^t.
\end{align} 
\end{corollary}


\section{Case studies: lower and upper bounds}
\label{sec-caseStudies}


The applicability of unlimited budget analysis is demonstrated through several case studies of RSH for maximising pseudo-Boolean functions.

\subsection{Instances of functions and algorithms used in case studies}
Three  pseudo-Boolean functions are considered in case studies. The first one is the family of linear functions, which was widely used in the theoretical study of RSH~\cite{he2001drift,lengler2015fixed,vinokurov2019fixed}.
\begin{eqnarray}
\textstyle f(x)= \sum^n_{i=1} c_i x_i,  \quad \mbox{where }   c_i >0. 
\end{eqnarray}
Its optimal solution   $x^*=(1\dots1)$ and  $f^*=\sum^n_{i=1} c_i$. An instance is the BinVal  function  $f(x)=\sum^n_{i=1}   2^i x_i$.

The second is the LeadingOnes function, which was an instance used in fixed budget performance~\cite{jansen2014performance}. 
\begin{align}
\textstyle f(x) =\sum^n_{i=1} \prod^i_{j=1} x_j,  \quad x \in \{0,1\}^n. 
\end{align} 
Its optimal solution $x^*=(1\cdots1)$ and  $f^*=n$. 

The third one is a multi-modal function, which was taken in runtime analysis of population-based EAs~\cite{he2002individual}.   Due to its zigzag shape (Fig.~1 in~\cite{he2002individual}), it is named Zigzag function.
\begin{equation} 
f(x) =
\left\{
\begin{array}{ll}
|x|, & \mbox{if $n-|x|$ is even}, \\
|x| - 2, & \mbox{\rm if $n-|x|$ is odd},
\end{array}
\right.
\end{equation}
where $|x|$ denotes the number of ones in $x$. Then its optimal solution $x^*=(1\cdots1)$ and $f^*=n$.

Three RSH algorithms are considered in this paper.  The first one is  random local search (RLS in short). 
 
\begin{itemize}
\item \textbf{Local search:} $y^{[t]} \leftarrow$ choose one bit of $x^{[t]}$ at random and flip it;
\item \textbf{Elitist selection}: $x^{[t+1]} \leftarrow$ select the best from
$y^{[t]} $ and $x^{[t]}$. 
\end{itemize}

The second  is (1+1) EA. 

\begin{itemize}
\item \textbf{Bitwise mutation}: $y^{[t]}\leftarrow$ flip each bit of $x^{[t]}$ with probability $\frac{1}{n}$; 
\item  \textbf{Elitist Selection}:  $x^{[t+1]}\leftarrow$ select the best one from $y^{[t]}$ and $x^{[t]}$.
\end{itemize}

The third algorithm is  simulated annealing  with a fixed temperature $T>0$ (SA-T in short). 

\begin{itemize}
\item \textbf{Neighbour search:} the neighbour of $x$ is the set of points $y$ with Hamming distance $d(x, y)=1$ or $2$. $y^{[t]}$is generated by with probability $1/2$, choosing one bit of $x^{[t]}$ at random and flipping it, otherwise choosing two bits of $x^{[t]}$ at random  and flipping them;
\item \textbf{Solution acceptance:} if $f(y^{[t]}) > f(x^{[t]})$, then accept $x^{[t+1]} \leftarrow y^{[t]}$; if $f(y^{[t]}) < f(x^{[t]})$,  then accept $x^{[t+1]} \leftarrow y^{[t]}$ with probability $\exp(-\frac{|f(x^{[t]})-f(y^{[t]})|}{T})$. 
    \item \textbf{Stopping criterion}: the algorithm halts if  an optimal solution is found.
\end{itemize} 
The above stopping criterion in SA-T is for the sake of analysis. Otherwise $e(x^{[t]})$ cannot be modelled by a super-martingale for any temperature $T>0$.

The three algorithms  can be modelled by Markov chains because the state of $x^{[t+1]}$ only depends on $x^{[t]}$. RLS and  (1+1) EA always can be modelled by supermartingales thanks to elitist selection, but SA-T can modelled by a supermartingale for  small $T$ but may not for large $T$.

\subsection{RLS  on linear functions}
\label{exaLinear}
For RLS on linear functions,  it is difficult to bound $f^{[t]}$ directly because the range of  coefficients  can be chosen arbitrarily large or small. However, it is simple to bound   $e^{[t]}$. 

We assume that  $x^{[t]}$ is a non-optimal solution such that   
 \begin{equation}
     x^{[t]}_i =\left\{\begin{array}{ll}
          1, & \mbox{if } i \in I,  \\
          0, &\mbox{otherwise}, 
     \end{array}
     \right.
 \end{equation}
 where $I \subset \{1, \cdots, n\}$ with $|I|<n$. The approximation error of $x^{[t]}$ is $e^{[t]}=\sum_{i \notin I} c_i $. The event of $e(x^{[t+1]}) < e(x^{[t]})$ happens if one bit  $x^{[t]}_j \notin I$ is flipped.  Its probability   is $1/n$.  
 
The average of error change (over all bits $j\notin I$) equals to
\begin{align}
 \textstyle   \Delta e(x^{[t]}) =  \sum_{j \notin I} c_{j}  \frac{1}{n}.
\end{align}
The ratio of error change equals to
\begin{align}
\frac{\Delta e(x^{[t]})}{e(x^{[t]})} = \frac{ \sum_{j \notin I} c_{j}  \frac{1}{n}}{\sum_{j \notin I} c_{j}}  =\frac{1}{n}.
\end{align}

Then we get   
\begin{align}
\label{equLinear-e1}
    e^{[t]} =& e^{[0]}   \left(1-\frac{1}{n}\right)^{  t},
\\
  \mbox{equivalently } 
\label{equLinear-f1}
  f^{[t]} =& f^*- e^{[0]} \left(1-\frac{1}{n}\right)^{  t}.
\end{align}

(\ref{equLinear-e1}) is an exact expression on $e^{[t]}$ for any $x^{[0]}$.  Surprisingly variant coefficients do not affect the formula. 

\begin{figure}[ht]
\centering
\includegraphics[width=0.5\columnwidth]{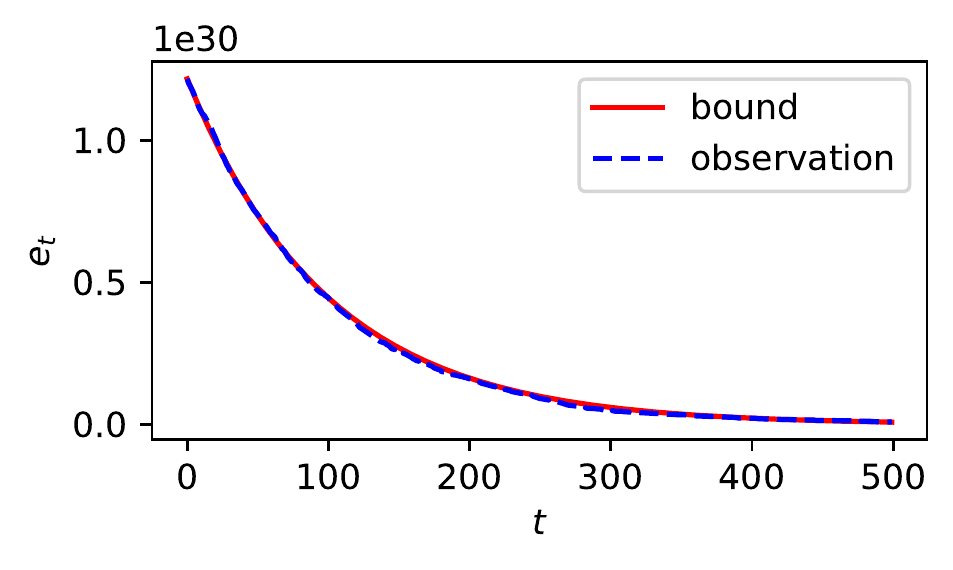} 
\caption{Observed mean $e^{[t]}$   with random initialisation on the BinVal function with $n=100$ and upper bound (\ref{equLinear-e1}). \label{fig-linear1}}
\end{figure}

We compare the derived exact formula (\ref{equLinear-e1}) with the experimental result on the BinVal function and present it in Fig.~\ref{fig-linear1}.

Since OneMax is a special case of linear functions, the above result generalises the work in \cite{jansen2014performance} from OneMax to all linear functions.

\subsection{(1+1) EA on linear functions}
Although  a few attempts have been made to analyse the fixed budget performance of   (1+1) EA on linear functions~\cite{lengler2015fixed,vinokurov2019fixed},     it is difficult to derive a general bound within fixed budget setting.  However, under the framework of unlimited budget analysis, it is  simple to derive a bound on $e^{[t]}$, then $f^{[t]}$. 

We assume that  $x^{[t]}$ is a non-optimal solution such that   
 \begin{equation}
     x^{[t]}_i =\left\{\begin{array}{ll}
          1, & \mbox{if } i \in I,  \\
          0, &\mbox{otherwise}, 
     \end{array}
     \right.
 \end{equation}
 where $I \subset \{1, \cdots, n\}$ with $|I|<n$. For  (1+1) EA, the event of $e(x^{[t+1]}) > e(x^{[t]})$ happens if one bit  $x^{[t]}_j \notin I$ is flipped and other bits are unchanged.  The probability of this event is $\frac{1}{n} (1-\frac{1}{n})^{n-1}$.  
 
The average of error change (over all bits $j\notin I$) satisfies
\begin{align}
  \textstyle  \Delta e(x^{[t]}) \ge   \sum_{j \notin I} c_{j}  \frac{1}{n}(1-\frac{1}{n})^{n-1}.
\end{align}
The ratio of error change satisfies
\begin{align}
\frac{\Delta e(x^{[t]})}{e(x^{[t]})} =  \frac{\sum_{j \notin I} c_{j}  \frac{1}{n}(1-\frac{1}{n})^{n-1}}{\sum_{j \notin I} c_{j}}  =\frac{1}{n}(1-\frac{1}{n})^{n-1}.
\end{align}

Then we get  
\begin{align}
\label{equLinear-e2}
    e^{[t]} \le& e^{[0]}   \left(1-\frac{1}{n}(1-\frac{1}{n})^{n-1}\right)^{  t}, 
\end{align}

\begin{figure}[ht]
\centering
\includegraphics[width=0.5\columnwidth]{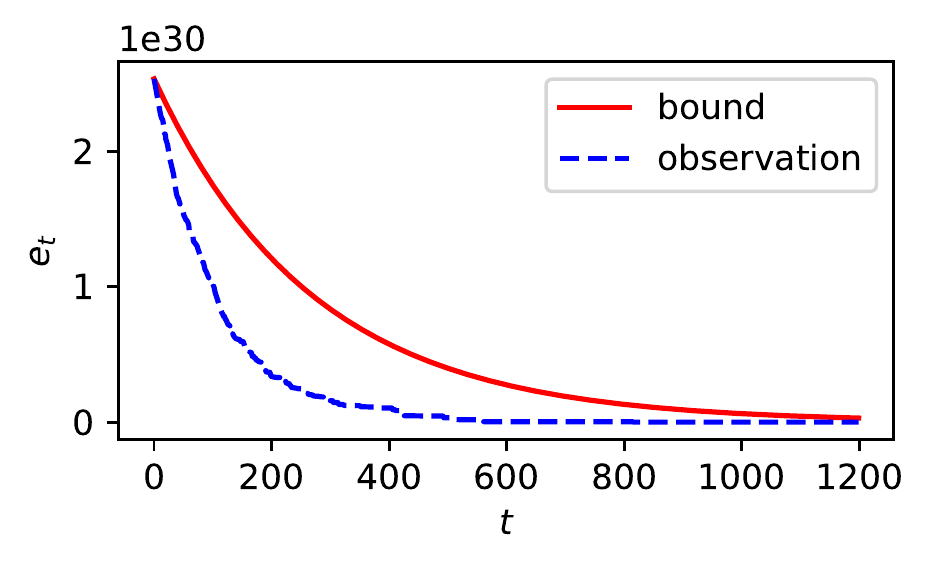} 
\caption{Observed mean $e^{[t]}$  of  (1+1) EA with random initialisation on the BinVal function   with $n=100$ and the upper bound (\ref{equLinear-e2}).
\label{fig-linear2}}
\end{figure}

(\ref{equLinear-e2}) is an upper bound on $e^{[t]}$ for any $x^{[0]}$ and  is reached at $|x^{[0]}|=n-1$. 
Again, we compare the derived upper bound (\ref{equLinear-e2}) on $e^{[t]}$ with the experimental result on  the BinVal function  and present it in Fig.~\ref{fig-linear2}.

Similar to the analysis of the upper bound on $e^{[t]}$, we derive a lower bound on $e^{[t]}$. We assume that  $x^{[t]}$ is a non-optimal solution such that   
 \begin{equation}
     x^{[t]}_i =\left\{\begin{array}{ll}
          1, & \mbox{if } i \in I,  \\
          0, &\mbox{otherwise}, 
     \end{array}
     \right.
 \end{equation}
 where $I \subset \{1, \cdots, n\}$ with $|I|<n$. Let $m$ denote $n-|I|$, the number of zeros. For  (1+1) EA, the event of $e(x^{[t+1]}) < e(x^{[t]})$ happens only if one of the following  mutually exclusive sub-events happens:
 \begin{enumerate}
     \item  one bit  $x^{[t]}_j \notin I$ is flipped and other bits $\notin I$ are unchanged. The probability of this event is at most $\frac{1}{n}(1-\frac{1}{n})^{m-1}$. The error is reduced by $c_{j}\frac{1}{n}(1-\frac{1}{n})^{m-2}$.
     \item two mutually different bits  $x^{[t]}_{j_1}, x^{[t]}_{j_2} \notin I$ are flipped and other bits $\notin I$ are unchanged.   The probability of this event is at most $\frac{1}{n^2}(1-\frac{1}{n})^{m-2}$. The error is reduced by $(c_{j_1}+c_{j_2})\frac{1}{n^2}(1-\frac{1}{n})^{m-2}$.
 
     \item $\cdots$ 
    \item all bits  $x^{[t]}_{j_1}, \cdots, x^{[t]}_{j_m} \notin I$ are flipped. The probability of this event is at most $\frac{1}{n^m}$. The error is reduced by $(c_{j_1}+\cdots+c_{j_m})\frac{1}{n^m}$.

 \end{enumerate}

The average of error change (over all bits $\notin I$) satisfies
\begin{align}
    \scriptstyle \Delta e(x^{[t]}) \le &\scriptstyle \sum_{j \notin I} c_{j}  \frac{1}{n} (1-\frac{1}{n})^{m-1}
    + \sum_{j_1 \neq j_2 \notin I} (c_{j_1}+c_{j_2})  \frac{1}{n}(1-\frac{1}{n})^{m-2}  \nonumber\\
    &+  \cdots + (c_{j_1}+\cdots+c_{j_m})  \frac{1}{n^m}. 
\end{align}
The ratio of error change satisfies
\begin{align}
\scriptstyle \frac{\Delta e(x^{[t]})}{e(x^{[t]})} \le & \scriptstyle \frac{\sum_{j \notin I} c_{j}  \frac{1}{n} (1-\frac{1}{n})^{m-1} +   +  \cdots + (c_{j_1}+\cdots+c_{j_m})  \frac{1}{n^m} }{ \sum_{j \notin I} c_{j}} \nonumber \\
 =& \scriptstyle \binom{m-1}{0}  \frac{1}{n} (1-\frac{1}{n})^{m-1}+   \cdots + \binom{m-1}{m-1} \frac{1}{n^m}
 =\frac{1}{n}.
\end{align}

Then we get   
\begin{align}
\label{equLinear-e3}
    e^{[t]} \ge& e^{[0]}   \left(1-\frac{1}{n}  \right)^{  t},
\end{align} 

\begin{figure}[ht]
\centering
\includegraphics[width=0.5\columnwidth]{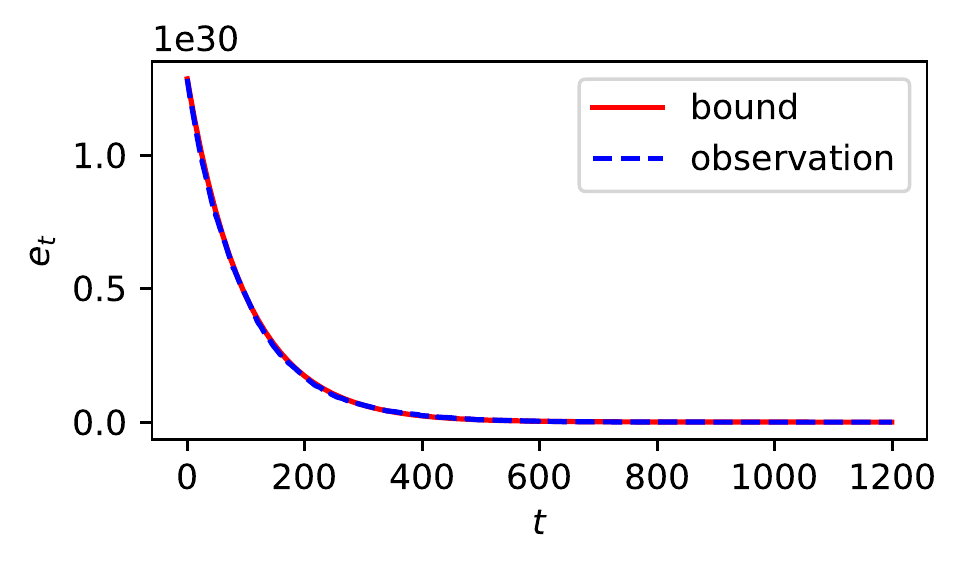} 
\caption{Observed mean $e^{[t]}$ of (1+1) EA with random initialisation on BinVal   with $n=100$ and the lower bound (\ref{equLinear-e3}).
\label{fig-linear3}}
\end{figure}

We compare the derived lower bound (\ref{equLinear-e3})   with the experimental result on the BinVal function and present it in Fig.~\ref{fig-linear3}.

Combining the upper bound (\ref{equLinear-e2}) and lower bound (\ref{equLinear-e3}) together, we obtain
\begin{align}
   e^{[0]}   \left(1-\frac{1}{n}  \right)^{  t} \le e^{[t]}\le e^{[0]}   \left(1-\frac{1}{n} (1-\frac{1}{n})^{n-1} \right)^{  t}. 
\end{align}

Compared with existing fixed budget analysis of linear functions~\cite{lengler2015fixed,vinokurov2019fixed}, unlimited budget analysis is simpler. 

\subsection{(1+1) EA on LeadingOnes function}
\label{exaLeadingOnes}
This example aims to show an advantage of unlimited budget performance over fixed budget performance, that is, a bound holds on any $t$.   
 (1+1) EA on LeadingOnes function has been analysed in~\cite{jansen2014performance} using fixed budget analysis. According to \cite[Theorem 13]{jansen2014performance}, a bound on $f^{[t]}$ is given as follows:   if $x^{[0]}$ is chosen uniformly at random, $t =(1-\beta)n^2 /\alpha(n)$ for any $\beta$ with $(1/2)+\beta'<\beta<1$ where $\beta'$ is a positive constant and $\alpha (n)=\omega(1)$, $\alpha (n)\ge 1$, then
\begin{align}
 \label{equLeading-Fixed}
  f^{[t]} &= 1+\frac{2t}{n} -o(\frac{t}{n}),\\
  \label{equLeading-Fixed-e}
  \mbox{equivalently } e^{[t]} &= n-1-\frac{2t}{n} +o(\frac{t}{n}).
\end{align} 

\begin{figure}[ht]
\centering
\includegraphics[width=0.5\columnwidth]{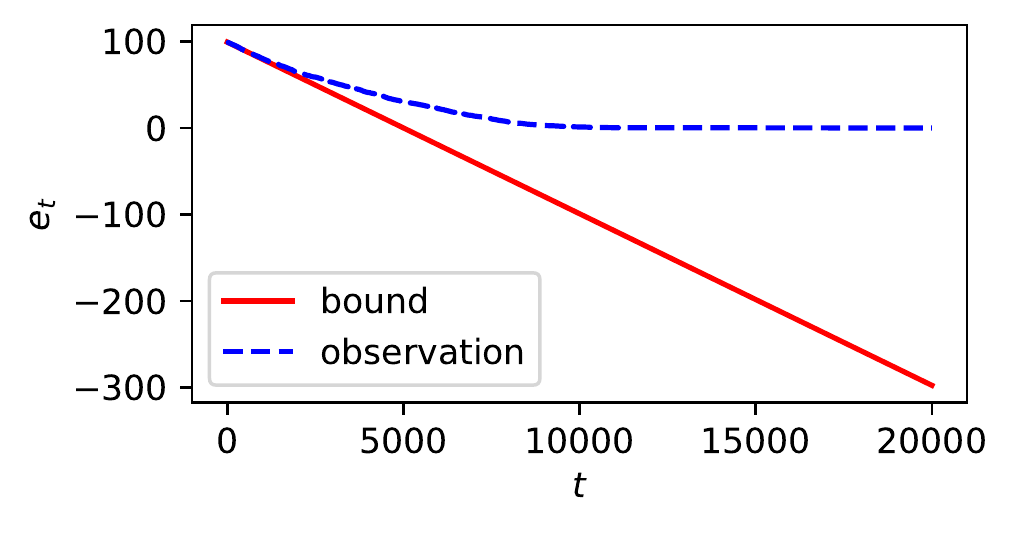} 
\caption{Observed mean $e^{[t]}$   of  (1+1) EA with random initialisation on LeadingOnes with $n=100$ and bound (\ref{equLeading-Fixed-e}) on $e^{[t]}$. \label{fig-leadingones-3}}
\end{figure}

(\ref{equLeading-Fixed-e}) is tight for  $t=o(t^2)$, but useless for large $t$. Fig.~\ref{fig-leadingones-3} depicts the observed mean $e^{[t]}$ and the bound (\ref{equLeading-Fixed-e}) on $e^{[t]}$. The bound is tight for small $t \le 1000$, but  useless for $t \ge 5000$ because the bound is negative. $e^{[t]}$ is always non-negative.  This is the limitation of fixed budget performance. 

With unlimited budget setting, we first find an upper bound on $e^{[t]}$ which holds for any $t$ and identical to (\ref{equLeading-Fixed-e}) if $t=o(n^2)$. 
Similar to  \cite[Theorem 13]{jansen2014performance}, we assume that the initial solution $x^{[0]}$ is chosen uniformly at random. Then we can draw a claim as follows.

Assume the initial solution $x^{[0]}$ is chosen uniformly at random, that is, $\Pr(x^{[0]}_i=1) =1/2$.  Then for any $t \ge 0$ and any $x^{[t]}=(1\cdots 1_i 0*\cdots *)$ where $i \in \{0, \cdots, n\}$ and  $* \in \{0, 1\}$ is a random variable, it holds $\Pr(*=1)= 1/2$.

We prove the claim by induction.  Because $x^{[0]}$ is chosen uniformly at random, we know $\Pr(x^{[0]}_i=1)=1/2$  for any $i \in \{1, \cdots, n\}$. We assume that the claim is true at some $t\ge 0$, that is,  $x^{[t]}=(1\cdots 1_i 0*\cdots *)$ where $i \in \{0, \cdots, n\}$ and $\Pr(x^{[t]}_j=1)= 1/2$ for any $j \ge i+1$. Now we let $x^{[t+1]}=(1\cdots 1_k 0*\cdots *)$. Thanks to elitist selection, it holds where $k \ge i$. For each bit $x^{[t]}_j$ such that $j\ge k+1$,  the change ($0\to 1$ or $1\to 0$) by bitwise mutation makes no contribution to the value of $f(x^{[t+1]})$. Thus 
\begin{align}
\scriptstyle\Pr(x^{[t+1]}_j=1)= \Pr(x^{[t+1]}_j=1\mid x^{[t]}_j=1) + \Pr(x^{[t+1]}_j=1\mid x^{[t]}_j=0)=\frac{1}{2}.
\end{align}
This means the claim is true at $t+1$. By induction, the claim is proven.  

We assume that  $x^{[t]}$ is a non-optimal solution such that   $x^{[t]}=(1\cdots 1_i0_{i+1}*\cdots*)$ for some $i \in \{0, \cdots,n-1\}$, where $\Pr(*=1) = 1/2$. 

\begin{itemize}
\item \textbf{\textbf Case 1: $i<n-1$.} We have 
\begin{equation} 
\scriptstyle \Pr(x^{[t+1]}=(1\cdots1_{i+1}0*\cdots *) \mid x^{[t]}) = \left(1-\frac{1}{n}\right)^{i}\frac{1}{n} \times \frac{1}{2}, 
\end{equation}
The formula is explained as follows. The first $i$ one-valued bits are unchanged with probability $(1-1/n)^i$. The $(i+1)$th zero-valued bit is flipped to one-valued with probability $1/n$. The flipping of bits labelled by $*$ dose not affect the fitness value.  From bitwise mutation, the $(i+1)$th bit satisfies $\Pr(x_{t+1,i+2}=0 \mid x^{[t]}_{i+2}=0)=1-1/n$ and    $\Pr(x_{t+1,i+2}=0 \mid x^{[t]}_{i+2}=1)=1/n$. Since $\Pr(x^{[t]}_{i+2}=*)=1/2$,  we get $\Pr(x_{t+1,i+2}=0 \mid x^{[t]}_{i+2}=*)=1/2$.

Following a similar argument, we draw that
\begin{align*}
&\scriptstyle \Pr(x^{[t+1]}=(1\cdots1_{i+2}0* \cdots *) \mid x^{[t]}) = \left(1-\frac{1}{n}\right)^{i} \frac{1}{n}  \times\frac{1}{2^2},\\
&\scriptstyle \Pr(x^{[t+1]}=(1\cdots1_{i+3}0* \cdots *) \mid x^{[t]}) = \left(1-\frac{1}{n}\right)^{i} \frac{1}{n}  \times\frac{1}{2^3}, \quad \cdots \cdots
 \end{align*}

The average of error  change is  
\begin{align} 
\label{equLO-error-change}
     \Delta e(x^{[t]})&\scriptstyle=\sum^{n-i}_{j=1}j \Pr(x^{[t+1]}=(1\cdots1_{i+j}0*\cdots *) \mid x^{[t]}) \nonumber \\
    & \scriptstyle=  \frac{1}{n} \left(1-\frac{1}{n}\right)^{i} \left ( \frac{1}{2}+\frac{2}{2^2}  +\cdots+\frac{n-i-1}{2^{n-i-1}} \right) =\Theta(\frac{1}{n}).
\end{align}
Then the ratio of error change is 
\begin{align}
\label{equCase1Ratio}
  \frac{\Delta e(x^{[t]})}{e(x^{[t]})}= &\scriptstyle\frac{1}{n} \left(1-\frac{1}{n}\right)^{i} \left ( \frac{1}{2}+\frac{2}{2^2}+ \cdots+\frac{n-i-1}{2^{n-i-1}} \right) \frac{1}{n-i}. 
\end{align}

Since for a large $n$ (say $n \ge 100$),  
\begin{align}
\label{equAppendix1}
 \scriptstyle \left(1-\frac{1}{n}\right)^{i} \left ( \frac{1}{2}+\frac{2}{2^2}+\frac{3}{2^3} +\cdots+\frac{n-i-1}{2^{n-i-1}} \right) \frac{1}{n-i} 
\ge   \left(2 -o(\frac{1}{n})\right) \frac{1}{n},
\end{align}
the ratio of error change is  lower-bounded by 
\begin{align}
\label{equCase1Ratio2}
\scriptstyle \frac{\Delta e(x^{[t]})}{e(x^{[t]})} \ge \left ( \frac{1}{2}+\frac{2}{2^2}+\frac{3}{2^3} +\cdots+\frac{n-1}{2^{n-1}} \right) \frac{1}{n^2} = \left(2 -o(\frac{1}{n})\right) \frac{1}{n^2}.
\end{align}

\item
\textbf{Case 2: $i=n-1$.} In this case, $x^{[t]}=(1\cdots 10)$. \begin{equation}
 \Pr(x^{[t+1]}=(1\cdots1) \mid x^{[t]}) = \left(1-\frac{1}{n}\right)^{n-1}\frac{1}{n}.
\end{equation}

The average of error  change is lower-bounded by
\begin{align}
     \Delta e(x^{[t]})= \left(1-\frac{1}{n}\right)^{n-1}\frac{1}{n}.
\end{align}
Then the ratio of error change is  
\begin{align}
\label{equCase2Ratio}
  \frac{\Delta e(x^{[t]})}{e(x^{[t]})}= \left(1-\frac{1}{n}\right)^{n-1}\frac{1}{n}.
\end{align}

\end{itemize} 

Summarising the two cases, we get  
\begin{align}
\frac{\Delta e(x^{[t]})}{e(x^{[t]})} \ge   \left(2 -o(\frac{1}{n})\right) \frac{1}{n^2}.
\end{align}
Then  we get an upper bound on the approximation error as 
\begin{align}
\label{equLeadingOne-e}
      e^{[t]}   \le e^{[0]} \left( 1- \frac{2}{n^2} + o(\frac{1}{n^3})\right)^t  ,
\end{align}

Now we estimate $e^{[0]}$. Because each bit in $x^{[0]}$ is set to 0 or 1 uniformly at random, we have for $i=0, \cdots, n$,
$$
 \Pr(x^{[0]}=(1\cdots1_{i}0*\cdots *))= \left(\frac{1}{2}\right)^{i+1}.
$$    Thus the initial fitness value  
\begin{align*}
 f^{[0]}=& \textstyle \sum^{n}_{i=1} i \Pr(x^{[0]}=(1\cdots1_{i}0*\cdots *)) \\
 =&\textstyle \sum^{n}_{i=1} i \left(\frac{1}{2}\right)^{i+1} =1 -o(\frac{1}{n}),
\end{align*}
and the initial error 
\begin{align} 
\label{equInitialError}
e^{[0]}= n- 1 +o(\frac{1}{n}).
\end{align}

\begin{figure}[ht]
\centering
\includegraphics[width=0.5\columnwidth]{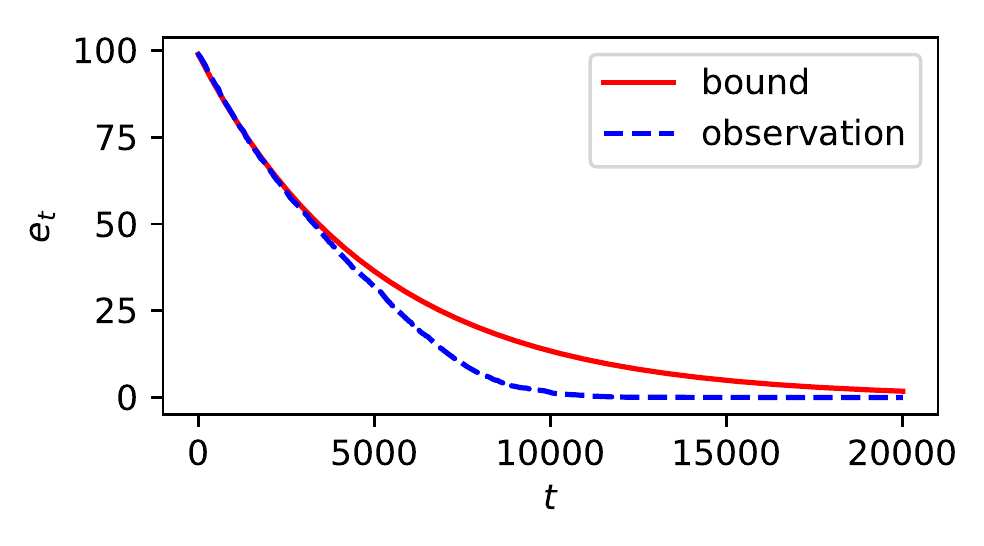} 
\caption{Observed mean $e^{[t]}$ of   (1+1) EA with random initialisation on LeadingOnes with $n=100$ and upper bound (\ref{equLeading-Unlimited}). \label{fig-leadingones}}
\end{figure}

Then  we get    
\begin{align} 
\label{equLeading-Unlimited}
    e^{[t]}  \le  \left( 1- \frac{2}{n^2} + o(\frac{1}{n^3})\right)^t  \left (n-1 +o(\frac{1}{n})\right).
\end{align} 
We compare the derived bound (\ref{equLeading-Unlimited})  with the  experimental result and present it in Figure~\ref{fig-leadingones}.
 
(\ref{equLeading-Unlimited}) is an exponetial function of $t$. When $t$ is small such that $t = o(n^2) $, it can be approximated by a linear function. This leads to (\ref{equLeading-Fixed-e}). The condition $t = o(n^2)$ implies $\frac{t^2}{n^4} = \frac{t}{n^2} o(1)$. From the binomial theorem,  we get
\begin{align}
\left( 1- \frac{2}{n^2} + o(\frac{1}{n^3})\right)^t &= 1-\frac{2t}{n^2} + \frac{2t}{n^2} o(1),
\end{align}
Then a linear approximation of (\ref{equLeading-Unlimited}) is given as
\begin{align}
\label{equLO-fixedbudget-upper}
      e^{[t]} &\le n-1-\frac{2t}{n} +o(\frac{t}{n})+o(\frac{1}{n}).
\end{align}
This upper bound is identical to (\ref{equLeading-Fixed-e}).

Next we estimate a  lower bound on $e^{[t]}$ for  (1+1) EA on LeadingOnes. We still assume that $x^{[0]}$ is chosen  uniformly at random. Let  $x^{[t]}$ be a non-optimal solution such that   $x^{[t]}=(1\cdots 1_i0_{i+1}*\cdots*)$ for some $i \in \{0, \cdots,n-1\}$ where $\Pr(*=1) = 1/2$. 
\begin{itemize}
\item \textbf{Case 1: $i<n-1$.}
The ratio of error change is given by (\ref{equCase1Ratio}) as
\begin{align} 
 \scriptstyle \frac{\Delta e(x^{[t]})}{e(x^{[t]})}= \frac{1}{n} \left(1-\frac{1}{n}\right)^{i} \left ( \frac{1}{2}+\frac{2}{2^2}+ \cdots+\frac{n-i-1}{2^{n-i-1}} \right) \frac{1}{n-i}. 
\end{align}

Since for a large $n$ (say $n \ge 100$),  
\begin{align}
\label{equAppendix2}
\scriptstyle\left(1-\frac{1}{n}\right)^{i} \left ( \frac{1}{2}+\frac{2}{2^2}+\frac{3}{2^3} +\cdots+\frac{n-i-1}{2^{n-i-1}} \right) \frac{1}{n-i} 
\le  \left(1-\frac{1}{n}\right)^{n-1}   ,
\end{align}
the ratio of error change is upper-bounded by 
\begin{align}
\frac{\Delta e(x^{[t]})}{e(x^{[t]})} \le \left(1-\frac{1}{n}\right)^{n-1} .
\end{align}

\item \textbf{Case 2: $i=n-1$.}  
The ratio of error change is given by (\ref{equCase2Ratio}) as 
\begin{align}
\label{equLO-lower-Case2}
  \frac{\Delta e(x^{[t]})}{e(x^{[t]})}= \left(1-\frac{1}{n}\right)^{n-1}\frac{1}{n}.
\end{align}
\end{itemize}

Summarising the two cases, we get  
\begin{align}
\label{equLO-lower-ratio}
\frac{\Delta e(x^{[t]})}{e(x^{[t]})} \le \left(1-\frac{1}{n}\right)^{n-1}\frac{1}{n} \approx \frac{1}{en}.
\end{align}
Then from (\ref{equInitialError}), we get  
\begin{align}
\label{equLeadingOne-e-lower}
      e^{[t]}   \ge&  \left (n-1 +o(\frac{1}{n})\right) \left( 1-  \frac{1}{en}\right)^t,
\end{align}

\begin{figure}[ht]
\centering
\includegraphics[width=0.5\columnwidth]{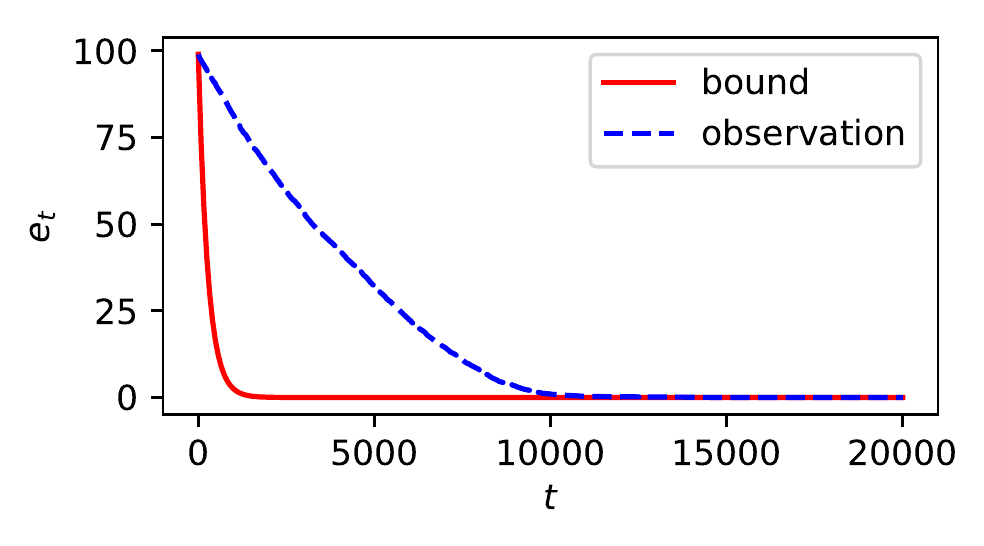} 
\caption{Observed mean $e^{[t]}$  of  (1+1) EA with random initialisation on LeadingOnes with $n=100$ and lower bound (\ref{equLeadingOne-e-lower}). \label{fig-leadingones-2}}
\end{figure}

The above  bounds  hold for any $t\ge 0$. 
compare the derived lower bound  (\ref{equLeadingOne-e-lower}) on $e^{[t]}$ with the experimental result  and present it in Figure~\ref{fig-leadingones-2}. The bound is tight for $t \ge 10000$ but not for $t \le 10000$. 

The above result reveals the limitation of Theorem~\ref{theoremBasic} which only gives a bound represented by an exponential function  as $c \lambda^t$.  But sometimes this expression is not good to approximate $e^{[t]}$. 
According to the theory of approximation analysis in~\cite{he2018theoretical}, an exact and general expression of $e^{[t]}$  is
\begin{align}\label{equGeneral}
 e^{[t]} =& \textstyle \sum_{i} \sum_{m}   c_{i_m} \binom{t}{l_{i,m}} (\lambda_{i})^{t-m+1}.
\end{align}  
A tight lower bound should approximate function (\ref{equGeneral}).  We will discuss this topic in a separate paper.

Our method  (Corollary~\ref{corollary}) can be used to evaluate the fixed budget performance of RSH too. For   (1+1) EA on LeadingOnes, we can draw the same bound on $e^{[t]}$ as  (\ref{equLeading-Fixed-e}) with a fixed budget $b = o(n^2)$. Recall that the upper bound on $e^{[t]}$ has been given by (\ref{equLO-fixedbudget-upper}) which is identical to (\ref{equLeading-Fixed-e}).

First we estimate the ratio of error change.  Define set  $\mathcal{S}_{o(n)}: =\{ x\mid x=(1\cdots 1_i 0 * \cdots * ), i =o(n) \}$. For any   $x=(1\cdots 1_i0_{i+1}*\cdots*) \in \mathcal{S}_{o(n)}$,
from (\ref{equCase1Ratio}), we know that the ratio of error change is 
\begin{align} 
 \scriptstyle 
 \frac{\Delta e(x)}{e(x)}= \frac{1}{n} \left(1-\frac{1}{n}\right)^{i} \left ( \frac{1}{2}+\frac{2}{2^2}+ \cdots+\frac{n-i-1}{2^{n-i-1}} \right) \frac{1}{n-i}. 
\end{align}

For a large $n$ (say $n \ge 100$) and $|x| =o(n)$,  
\begin{align}
  \frac{\Delta e(x)}{e(x)}  &= \frac{1}{n} \left(2 -o(\frac{1}{n})\right) \left(\frac{1}{n} +o(\frac{1}{n})\right)\nonumber \\
  &=\frac{2}{n^2} - o(\frac{1}{n^2}).
\label{equLO-lower-ratio2}
\end{align}

Next we prove a claim, that is, within a fixed budget $t\le b=o(n^2)$, the probability 
$
\Pr(x^{[t]} \in \mathcal{S}_{o(n)}) \ge 1-o(1).
$

Let $a(n) =n^2/b$. Since $b=o(n^2)$, we have 
\begin{align*}
\lim_{n \to \infty}  a(n)=+\infty,  
\lim_{n \to \infty} \ln a(n)=+\infty,  
\frac{\ln a(n)}{ a(n)}=o(1).
\end{align*}

From (\ref{equLO-error-change}), the error change satisfies $\Delta e(x^{[t]}) =\Theta(1/n)$, then we have
\begin{align}
    \mathbb{E}[f(x^{[t]})]  \le  \Theta(\frac{b}{n}).
\end{align}
According to Markov inequality, we get
\begin{align*}
   \Pr(f(x^{[t]})  \ge \frac{n}{\ln a(n)}) \le& \frac{\mathbb{E}[f(x^{[t]})]}{\frac{n}{\ln a(n)}}  \le  \Theta(\frac{b}{n}) \frac{\ln a(n)}{n} \\    
   =&\frac{\ln a(n)}{ a(n)} \Theta(1)=o(1). 
\end{align*}

Then the probability $\Pr(f(x^{[t]}) < \frac{n}{\ln a(n)})=1-o(1)$. Because $\frac{n}{\ln a(n)}=o(n)$, we get $\Pr(x^{[t]}\in\mathcal{S}_{o(n)}) = 1-o(1)$. Thus, with probability $1-o(1)$, $x^{[0]}, \cdots, x^{[b]} \in \mathcal{S}_{o(n)}$ and the ratio of error change satisfies (\ref{equLO-lower-ratio2}). 
Then with probability $1-o(1)$, we have a lower bound on $e^{[t]}$  as  
\begin{align*} 
     e^{[t]}   =  e^{[0]} \left( 1- \frac{2}{n^2} + o(\frac{1}{n^2})\right)^t,
 \end{align*} 
where $e^{[0]} = \left (n-1 +o(\frac{1}{n})\right)$.
 
Since $t \le b = o(n^2) $,   we have $\frac{t^2}{n^4} = \frac{t}{n^2} o(1)$. From the binomial theorem,  we have a linear approximation of $e^{[t]}$ as
\begin{align} 
\label{equLO-fixed}
      e^{[t]} &= e^{[0]} \left( 1- \frac{2t}{n^2} + o(\frac{2t}{n^2})\right),\nonumber\\
       &= n-1- \frac{2t}{n}+  o(\frac{t}{n}).
\end{align}  

This  expression is identical to (\ref{equLeading-Fixed-e}) or equivalent to (\ref{equLeading-Fixed}) from fixed budget performance~\cite{jansen2014performance}. This study shows that within fixed budget setting, Corollary~\ref{corollary} can be used to derive a tighter bound on $e^{[t]}$ than Theorem~\ref{theoremBasic}.

\section{Case studies: algorithm comparison}
\label{sec-comparison}
Like runtime analysis, unlimited budget analysis can be used to compare the performance of one  RSH algorithm with different parameter setting or  two different RSH algorithms.  The comparison is based on Theorem~\ref{theoremOneStep} which states $e^{[t]}\le e^{[0]} (1-\delta_{\min})^t$. Given two algorithms, we compare  their $\delta_{\min}$ values and the upper bound on $e^{[t]}$. 

\subsection{Comparison of (1+1) EA with different mutation rates}
In order to achieve the best performance, it is common practice to fine-tune some parameter in RSH. Consider the bitwise mutation rate, that is to flip each bit with probability $p$. This example aims to investigate the best rate $p$ of  (1+1) EA on linear functions  in terms of the upper bound on  $e^{[t]}$.  

We assume that  $x^{[t]}$ is a non-optimal solution such that   
 \begin{equation}
     x^{[t]}_i =\left\{\begin{array}{ll}
          1, & \mbox{if } i \in I,  \\
          0, &\mbox{otherwise}, 
     \end{array}
     \right.
 \end{equation}
 where $I$ is a subset of $\{1, \cdots, n\}$ with $|I|<n$. $m$ denotes $n-|I|$, the number of zeros. For   (1+1) EA, the event of $f(x^{[t+1]}) > f(x^{[t]})$ happens  if one of the following mutually exclusive sub-events happens:
 \begin{enumerate}
     \item  one bit  $x^{[t]}_j \notin I$ is flipped and other bits are unchanged. The probability of this event is at most $p(1-p)^{n-1}$. The error is reduced by $c_{j}p(1-p)^{n-1}$.
     \item two mutually different bits  $x^{[t]}_{j_1}, x^{[t]}_{j_2} \notin I$ are flipped and other bits $\notin I$ are unchanged.   The probability of this event is at most $p^2(1-p)^{n-2}$. The error is reduced by $(c_{j_1}+c_{j_2})p^2(1-p)^{n-2}$.
     \item $\cdots$ 
    \item all bits  $x^{[t]}_{j_1}, \cdots, x^{[t]}_{j_m} \notin I$ are flipped. The probability of this event is at most $p^{n-m}$. The error is reduced by $(c_{j_1}+\cdots+c_{j_m})p^{n-m}$.
 \end{enumerate}

The average of error change (over all bits $\notin I$) satisfies
\begin{align}
    \Delta e(x^{[t]}) \ge &\scriptstyle \sum_{j \notin I} c_{j}  p (1-p)^{n-1} + \sum_{j_1 \neq j_2 \notin I} (c_{j_1}+c_{j_2})  p^2(1-p)^{n-2} \nonumber \\
    &\scriptsize+  \cdots + (c_{j_1}+\cdots+c_{j_m})  p^{m}(1-p)^{n-m}. 
\end{align}
The ratio of error change satisfies
\begin{align}
\frac{\Delta e(x^{[t]})}{e(x^{[t]})} \ge   &\scriptstyle \frac{\sum_{j \notin I} c_{j}  p (1-p)^{n-1}    +  \cdots + (c_{j_1}+\cdots+c_{j_m})  p^{m}(1-p)^{n-m}}{ \sum_{j \notin I} c_{j}} \nonumber \\
 = &\scriptstyle \binom{m-1}{0}  p (1-p)^{n-1}   +\cdots + \binom{m-1}{m-1} p^m (1-p)^{n-m}\nonumber\\
 =&p(1-p)^{n-m} \ge p(1-p)^{n-1}.
\end{align}

The above inequality is reached at $|x^{[t]}|=n-1$. When $|x^{[t]}|=1$, it means $x^{[t]}$ includes only one zero-valued bit.   The event of $f(x^{[t+1]})> f(x^{[t]})$ happens if and only if this unique zero-valued bit in $x^{[t]}$ is flipped and other bits are unchanged. The probability of this event equals to $  p (1-p)^{n-1}$. In the ratio of error change equals to
\begin{align}
\frac{\Delta e(x^{[t]})}{e(x^{[t]})} = \frac{ p (1-p)^{n-1}}{1} =p(1-p)^{n-1}.
\end{align}

Then we get  the approximation error as
\begin{align}
\label{equLinear-e4}
    e^{[t]} \le e^{[0]}   \left(1-p(1-p)^{n-1}  \right)^{  t}.
\end{align}

Thus the minimal ratio of error change is
\begin{align*}
    \delta_{\min}(p)  =p(1-p)^{n-1}.
\end{align*}

Then we get an upper bound on the approximation error as
\begin{align}
\label{equOneMax-e2}
    e^{[t]} \le e^{[0]} (1-\delta_{\min}(p))^t= e^{[0]} (1-p(1-p)^{n-1})^t.
\end{align}

Now we find the value of $p$ of minimising the upper bound $(1-p(1-p)^{n-1})^t$. This is equivalent to
\begin{align} 
    \max    \delta_{\min}(p)  = p(1-p)^{n-1}, \quad p \in (0,1). 
\end{align}
For $p \in (0,1)$, we know $\delta_{\min}(p)$ takes the maximal value at $p={1}/{n}$. With the value, the upper bound on $e^{[t]}$ is smallest.
 
\begin{figure}[ht]
\centering
\includegraphics[width=0.75\columnwidth]{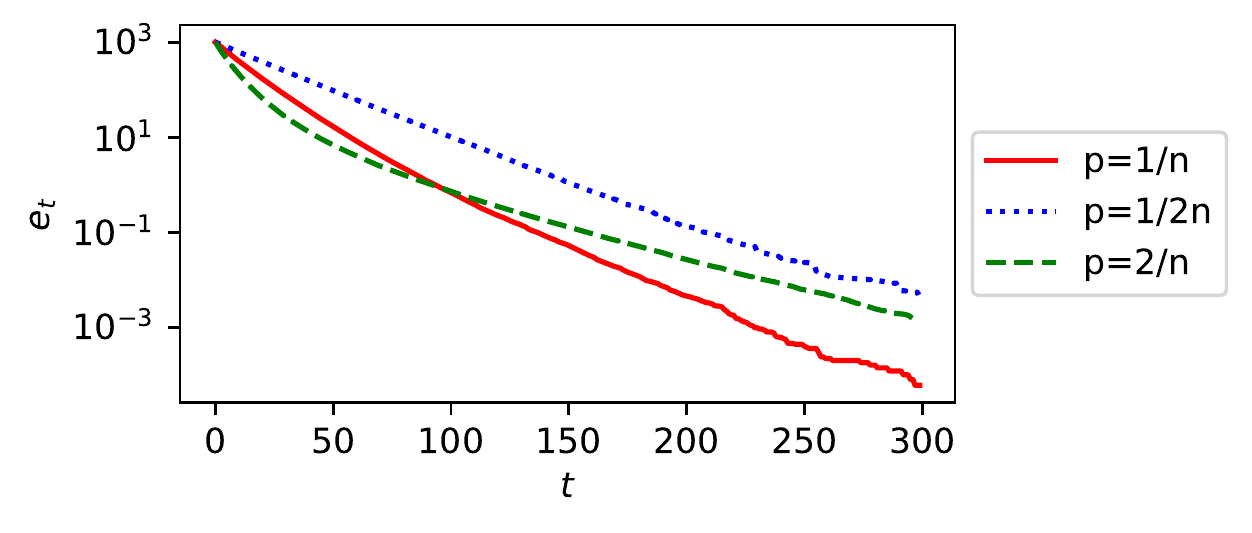} 
\caption{Observed mean $e^{[t]}$ of  (1+1) EA with random initialisation on  BinVal with $n=10$ for  $p=1/n, 2/n, 1/(2n)$ respectively.  \label{fig-onemax2}}
\end{figure}

Fig.~\ref{fig-onemax2} shows the observed value of $e^{[t]}$  in computational experiments of  (1+1) EA on the BinVal function with mutation rates $p=1/n, 2/n, 1/(2n)$ respectively. Experimental results reveals that $p=1/n$ is the best among  three mutation rates in terms of the approximation error. The figure also shows that $p=2/n$ is  better than other two at the beginning of search. This result could be rigorously proven using Corollary~\ref{corollary}.

\subsection{Comparison of (1+1) EA and SA-T on Zigzag}
In order to show a RSH algorithm is better than another on an optimisation problem, it is  common practice to compare $e^{[t]}$ or $f^{[t]}$  achieved by the  two RSH algorithms.   This example aims to compare the upper bound on $e^{[t]}$ of   (1+1) EA and SA-T on the Zigzag function. 

First we analyse   (1+1) EA. We assume that  $x^{[t]}$ is a non-optimal solution such that  $|x^{[t]}|=i <n$. We estimate the minimum ratio  of error change $\delta_{\min}$. Obviously it is sufficient to consider even numbers $i$ because the ratio of error change at an odd number $j$ is that at its neighbour even number $j+1$ or $j-1$.

Given an even number $i <n$, $e(x^{[t+1]}) < e^{[t]}$ happens if two 0-valued bits are flipped and other are unchanged. The probability of this event is $\binom{n-i}{2} \frac{1}{n^2} (1-\frac{1}{n})^{n-2}$. The error change satisfies
\begin{align}
    \Delta e(x^{[t]}) \ge \binom{n-i}{2} \frac{1}{n^2} (1-\frac{1}{n})^{n-2} \times 2.
\end{align}
The ratio of error change is lower-bounded by
\begin{align}
   \scriptstyle \frac{\Delta e(x^{[t]})}{e(x^{[t]})} \ge  \frac{(n-i)(n-i-1)}{n^2} (1-\frac{1}{n})^{n-2}  \times \frac{1}{n-i} \ge \frac{1}{n^2} (1-\frac{1}{n})^{n-2}.
\end{align}
The above lower-bound is reached if $|x^{[t]}|=n-2$. Then 
\begin{align}
&\delta_{\min} = \frac{1}{n^2} (1-\frac{1}{n})^{n-2} \approx \frac{1}{en^2} .\\
\label{equEAUpperBound}
    &e^{[t]}   \approx e^{[0]}(1-\frac{1}{en^2}  )^t.
\end{align}

Next   we analyse SA-T. We assume that  $x^{[t]}$ is a non-optimal solution such that  $|x^{[t]}|=i <n$. We estimate the minimum ratio  of error change $\delta_{\min}$. Obviously it is sufficient to consider even numbers $i$ because the ratio of error change at an odd number $j$ is that at its neighbour even number $j-1$ or $j+1$.

Given an even number $i <n$, the event $e(x^{[t+1]}) \neq e(x^{[t]})$ happens if and only if one of the following  events happens.
\begin{enumerate}
    \item \textbf{$|x^{[t+1]}|=i+2$.} This event happens if two 0-valued bits are flipped and other are unchanged.  The probability of this event is $\binom{n-i}{2} \frac{1}{n^2}\frac{1}{2}$. The error change is positive,
    \begin{align}
        \Delta e(x^{[t]}) = \frac{(n-i)(n-i-1) }{2n^2}.
    \end{align}
    \item \textbf{$|x^{[t+1]}|=i+1$.} This event happens if one 0-valued bit is flipped and the child is accepted.  Its probability  is $\binom{n-i}{1} \frac{1}{n} \frac{1}{2} \exp(-\frac{1}{T})$.  The error change is negative,
    \begin{align}
        \Delta e(x^{[t]}) =  -\frac{(n-i)}{2 n^2} \exp(-\frac{1}{T}).
    \end{align}
    
    \item \textbf{$|x^{[t+1]}|=i-1$.} This event happens if one 1-valued bit is flipped and the child is accepted.   Its probability is $\binom{i}{1} \frac{1}{n} \frac{1}{2} \exp(-\frac{3}{T})$.  The error change is negative,
    \begin{align}
        \Delta e(x^{[t]}) =  -\frac{i}{2 n^2} \exp(-\frac{3}{T}).
    \end{align}
    
    \item \textbf{$|x^{[t+1]}|=i-2$.} This event happens if two 1-valued bits are flipped and the offspring is accepted.   Its probability is $\binom{i}{2} \frac{1}{n} \frac{1}{2} \exp(-\frac{2}{T})$.  The error change is negative,
    \begin{align}
        \Delta e(x^{[t]}) =  -\frac{i(i-1)}{2 n^2} \exp(-\frac{2}{T}).
    \end{align}
\end{enumerate}

The total error change is 
\begin{align} 
\Delta e(x^{[t]}) = &\frac{(n-i)(n-i-1) }{2n^2} - \frac{(n-i)}{2 n^2} \exp(-\frac{1}{T}) \nonumber\\
& -\frac{i}{2 n^2} \exp(-\frac{3}{T})-\frac{i(i-1)}{2 n^2} \exp(-\frac{2}{T}).
\end{align}

The ratio of error change is
\begin{align}
\label{equErrorChange}
\scriptstyle \frac{\Delta e(x^{[t]})}{e(x^{[t]})} = &\scriptstyle \frac{n-i-1 }{2n^2} - \frac{1}{2 n^2} \exp(-\frac{1}{T}) \nonumber\\
&\scriptstyle  -\frac{i}{2(n-i) n^2} \exp(-\frac{3}{T})-\frac{i(i-1)}{2(n-i) n^2} \exp(-\frac{2}{T}).
\end{align}

We choose temperature $T$ sufficiently small so that the last three negative items in~(\ref{equErrorChange}) is greater than $-o(\frac{1}{n^2})$. Then  
\begin{align} 
\frac{\Delta e(x^{[t]})}{e(x^{[t]})}  \ge \frac{n-i-1 }{2n^2}  -o(\frac{1}{n^2})  \ge   \frac{1}{2n^2} -  o(\frac{1}{n^2}) .
\end{align}
The above inequality is reached at $|x^{[t]}|=n-2$ and $\delta_{\min}=\frac{1}{2n^2} -  o(\frac{1}{n^2}).$ 
The  error is upper-bounded by 
\begin{align}
\label{equSAUpperBound}
    e^{[t]} \le e^{[0]}(1-\frac{1}{2n^2} + o(\frac{1}{n^2}))^t.
\end{align}

Comparing (\ref{equEAUpperBound}) with (\ref{equSAUpperBound}), we see that upper bound on $e^{[t]}$  of  (1+1) EA  is slightly  larger than that of SA-T.

\begin{figure}[ht]
\centering
\includegraphics[width=0.75\columnwidth]{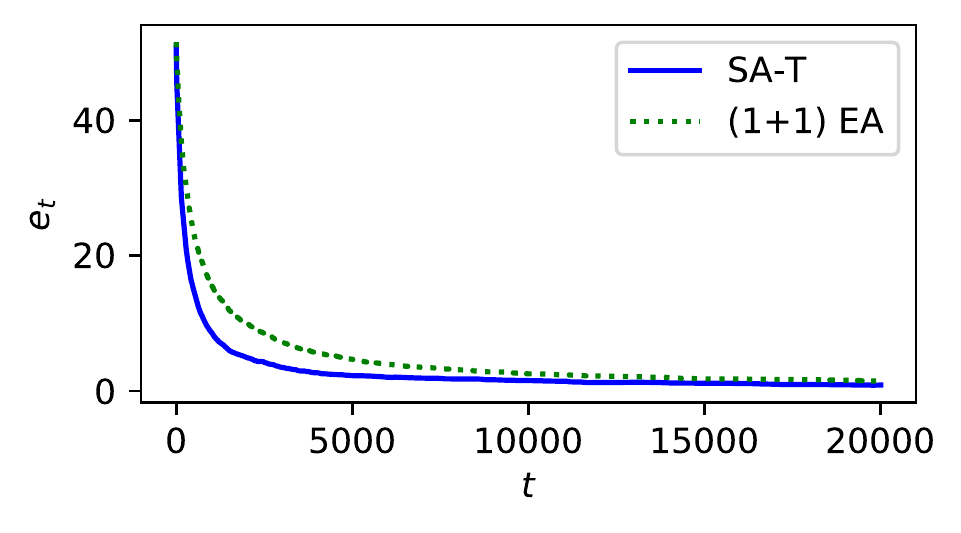} 
\caption{Observed mean $e^{[t]}$ of  (1+1) EA and SA-T with random initialisation on  Zigzag functions respectively, where $n=100$.  \label{fig-error-compare}}
\end{figure}

Fig.~\ref{fig-error-compare} shows the observed mean  $e^{[t]}$  in computational experiments of   (1+1) EA and SA-T. The error $e^{[t]}$ of (1+1) EA is slightly larger than that of SA-T.

\section{Conclusions}
\label{sec-conclusions}

This paper presents a new approach, called   unlimited budget analysis, for evaluating the   performance of RSH measured by expected function values or approximation error after an arbitrary number of computational steps. Its novelty is to first derive a bound on the approximation error, then a bound on the fitness value.  The approach reveals that the upper and lower bounds on the approximation error or fitness value can be estimated by the ratio of error change in one or multiple steps.

The applicability of the new approach is demonstrated  by several case studies.  For random local search and  (1+1)~EA on linear functions, they are difficult to existing methods in fixed budget setting,  but using unlimited budget analysis have derived general bounds on all linear functions. For (1+1)  EA on LeadingOnes,  
 unlimited budget analysis extends the results obtained by fixed budget analysis from a fixed number of computational  steps to an arbitrary number of steps. 
For  (1+1) EA on linear functions, the best bitwise mutation rate  is identified as $1/n$ in terms of   the approximation error. It is also found that the performance of  (1+1) EA is slightly worse than simulated annealing  on the Zigzag function. 

About future research, one is to consider the ratio of error change in multiple steps to see how much bounds can be strengthened. Another is to apply unlimited budget analysis to more problems and  algorithms.



\section*{supplement: Proof of inequalities (\ref{equAppendix1}) and (\ref{equAppendix2})}
In the supplement, we provide the proof of (\ref{equAppendix1}) and (\ref{equAppendix2}) with detail. The proof of the two inequalities is purely based on mathematics  but not related to randomised search heuristics. 

Denote
\begin{align}
  g(i) = \textstyle\left(1-\frac{1}{n}\right)^{i} \left ( \frac{1}{2}+\frac{2}{2^2}+\frac{3}{2^3} +\cdots+\frac{n-i-1}{2^{n-i-1}} \right)\frac{1}{n-i},
\end{align} 
we want to prove
\begin{align}
    \frac{2}{n}\le  g(i) \le \left(1-\frac{1}{n}\right)^{n-1}.
\end{align} 

Denote
\begin{align}
  g_a(i) &= \textstyle\left(1-\frac{1}{n}\right)^{i}  \frac{1}{n-i}.\\
  g_b(i) &= \left ( \frac{1}{2}+\frac{2}{2^2}+\frac{3}{2^3} +\cdots+\frac{n-i-1}{2^{n-i-1}} \right).
\end{align} 
For $0\le i \le n-2$, $g_a(i)$ is a monotonically  increasing function of $i$. $g_b(i)$ is a monotonically decreasing function of $i$ and
$$
\lim_{n \to +\infty} g_b(0)=2.
$$

First we prove $g(i) \le \left(1-\frac{1}{n}\right)^{n-1}$.

\textbf{Case 1: $i=n-2,n-3,n-4$.} For sufficient large $n$ (e.g. $n\ge 100$), we have
\begin{align*}
    g(n-2)&= (1-\frac{1}{n})^{n-2} \frac{1}{2} \times \frac{1}{2} \le \left(1-\frac{1}{n}\right)^{n-1}.\\
    g(n-3)&= (1-\frac{1}{n})^{n-3} (\frac{1}{2}+\frac{2}{2^2}) \times \frac{1}{3} \le \left(1-\frac{1}{n}\right)^{n-1}.\\
    g(n-4)&= (1-\frac{1}{n})^{n-4} (\frac{1}{2}+\frac{2}{2^2}+\frac{3}{2^3}) \times \frac{1}{4} \le \left(1-\frac{1}{n}\right)^{n-1}.
\end{align*}

\textbf{Case 2: $0\le i \le n-4$.}  Since
\begin{align*}
    g_b(i) \le 2 .
\end{align*}
then for sufficient large $n$ (e.g. $n\ge 100$), we have
\begin{align}
    g(i) &\le 2 g_a(i) \le 2 g_a(n-4) \nonumber \\
    &=2\times \frac{1}{4} (1-\frac{1}{n})^{n-4} < \left(1-\frac{1}{n}\right)^{n-1}.
\end{align}

Combining the above two cases, we finish the proof.  

Next we prove $g(i) \ge (2-o(\frac{1}{n})) \frac{1}{n}$.

\textbf{Case 1: $i\le (1-\frac{1}{4e})n$.} We have
\begin{align}
    g_b(i) \ge 2-\Theta(\frac{\frac{n}{4e}}{2^{\frac{n}{4e}}})=2-o(\frac{1}{n}).
\end{align}
Then we get 
\begin{align}
    g(i) &\ge (2-o(\frac{1}{n})) g_a(i) \ge (2-o(\frac{1}{n})) g_a(0)\nonumber\\
    & =  (2-o(\frac{1}{n})) \frac{1}{n}.
\end{align}

\textbf{Case 2: $i \ge (1-\frac{1}{4e})n$.} We have
\begin{align}
    g_a(i) \ge \frac{1}{2}.
\end{align} 
Then we get 
\begin{align}
    g(i) &\ge \frac{1}{2} g_a(i) \ge \frac{1}{2} g_a(n-\frac{n}{4e})\nonumber\\
    & =\frac{1}{2} (1-\frac{1}{n})^{n-\frac{n}{4e}} \frac{4e}{n} \ge \frac{2}{n}.
\end{align}

Combining the above two cases, we finish the proof.  

\end{document}